\documentclass[twoside,11pt]{article}

\usepackage{blindtext}
\usepackage[preprint]{jmlr2e}
\usepackage{amsmath}
\usepackage{amsfonts}
\usepackage{bm}
\usepackage{minted}

\usepackage{lastpage}
\jmlrheading{23}{2022}{1-\pageref{LastPage}}{1/21; Revised 5/22}{9/22}{21-0000}{Alex Shtoff}

\newminted[pycode]{python}{tabsize=4, fontsize=\scriptsize, frame=lines, framesep=\fboxsep, rulecolor=\color{gray!40}}

\def\eqref#1{equation~\ref{#1}}

\def\1{\bm{1}}

\def\vc{{\bm{c}}}

\def\vv{{\bm{v}}}

\def\vx{{\bm{x}}}

\DeclareMathAlphabet{\mathsfit}{\encodingdefault}{\sfdefault}{m}{sl}
\SetMathAlphabet{\mathsfit}{bold}{\encodingdefault}{\sfdefault}{bx}{n}

\ShortHeadings{TorchCurves: Differentiable Parametric Curves}{TorchCurves: Differentiable Parametric Curves}
\firstpageno{1}

\begin{document}

\title{TorchCruves: Differentiable Parametric Curves}

\author{\name Alex Shtoff \email alex.shtf@gmail.com}


\maketitle

\begin{abstract}
We present TorchCurves, a PyTorch library that treats vector-valued parametric curves as learnable primitives. Its central operation maps batches of scalar inputs to points on curves represented in B-spline or Legendre polynomial bases, with differentiation through both inputs and coefficients. Curve coefficients can be optimized jointly with the surrounding model or produced by another network. The library provides trainable modules, reusable basis modules, and a functional interface, supporting applications ranging from continuous feature embeddings to Kolmogorov–Arnold networks and functional regression. Its implementation combines classical evaluation algorithms with vectorized PyTorch operations, emphasizing readability and maintainability without custom hardware kernels. Timing experiments compare curve evaluation against NumPy and SciPy and KAN training against EfficientKAN across a range of problem sizes. The results illustrate where this generality remains computationally practical and where overhead is substantial. Library is available on GitHub at \url{https://github.com/alexshtf/torchcurves}.
\end{abstract}


\section{Introduction}

\begin{figure}[htbp]
    \centering
    \includegraphics[width=0.75\linewidth]{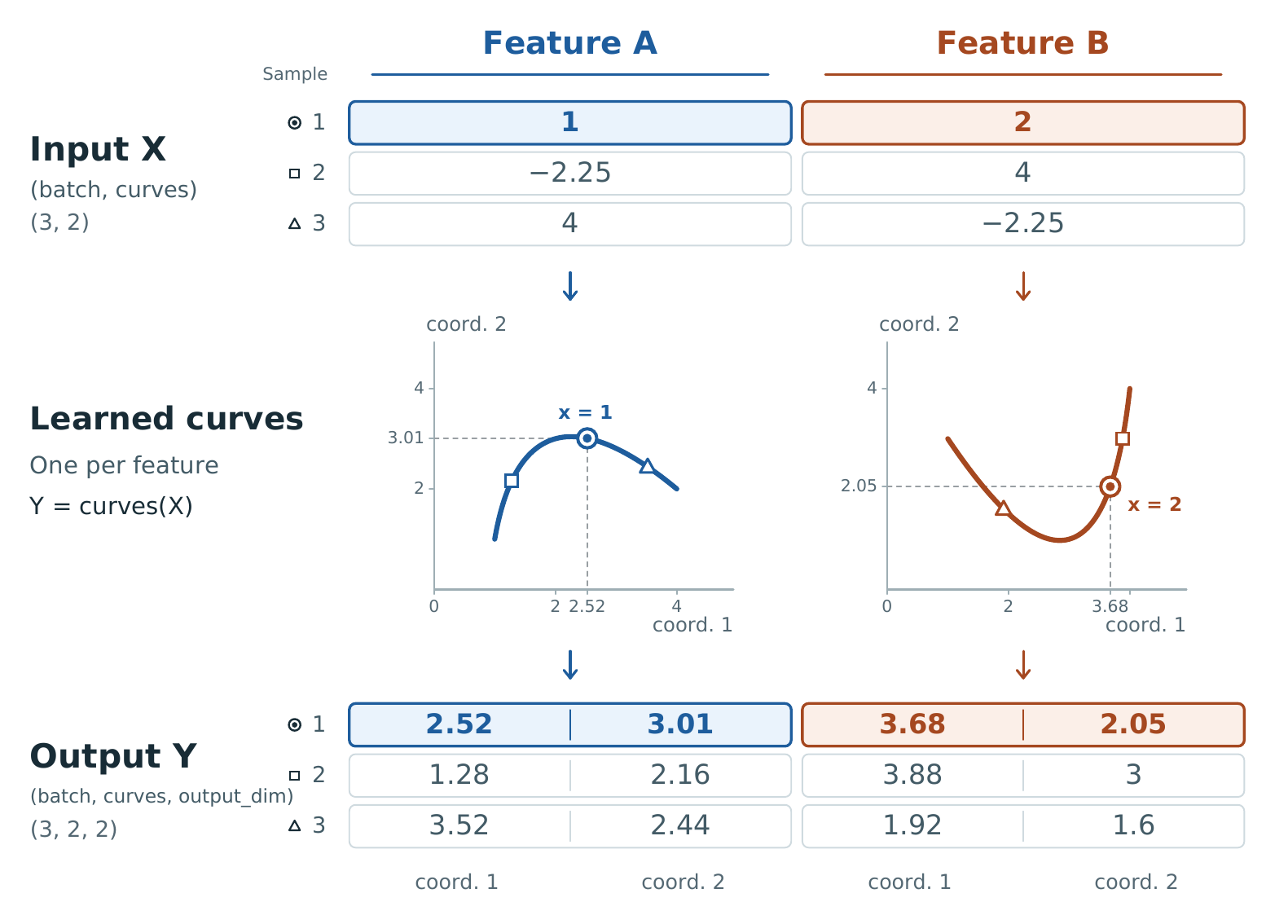}
    \caption{The basic operation behind TorchCurves. Input: a mini-batch of numerical features of size. Output: a mini-batch of points on multi-dimensional curves. }
    \label{fig:basic_op}
\end{figure}

The use of parametric curves spans a wide variety of scientific and engineing disciplines, both in machine learning and outside of it. Examples include spline regression, continuous features embeddings \citep{shtoff2024functionbasis}, Kolmogorov-Anrnold networks \citep{liu2025kan}, computer aided design \citep{bezier1968renault} and others. Yet, there seems to be no software library providing a common, composable interface for curves as learnable primitives, but rather every re-implements its own high-level object directly. TorchCurves puts the learnable parametric curve as a primitive in its own right, and facilitates its use in a variety of model achitectures. Its genericity allows PyTorch \citep{ansel2024pytorch} users to build models that include learnable parametric curves as a component.

The essence of TorchCurves is \emph{one operation} - transforming a mini-batch of input features to a mini-batch of points on curves in $\mathbb{R}^d$, as depicted in Figure \ref{fig:basic_op}.  At this stage we support B-Spline and Legendre polynomials. Our implementation is vectorized and there are no loops on the feature, batch and curve dimensions. But we do not have custom kernels for specialized hardware - we chose to implement everything in pure PyTorch for readability and maintainability reasons. This basic operation is differentiable in both its parameters, the input features and the curve parameters, allowing both learning a curve and back-propagating to previous network layers.

The next sections are structures as follows: in Section \ref{sec:applications} we present the main components of the library and present examples of applications: continuous feature embeddings, Kolmogorov-Arnold networks, and functional regression. In Section \ref{sec:impl} we discuss some implementation and testing details.

\section{Related work}

Polynomial and spline representations are well established in numerical software. NumPy \citep{numpy2020harris} and SciPy \citep{scipy2020pauli} provide routines for representing, evaluating, and differentiating such functions. Differentiable spline packages also exist within the PyTorch ecosystem: \texttt{torchcubicspline} \citep{kidgertorchcubicspline} constructs natural cubic interpolants from observations, while \texttt{torch-bspline} \citep{melchers2025torchbspline} provides differentiable B-spline evaluation.

In machine learning, mappings from scalar values to learned vectors appear in numerical feature embeddings \citep{gorishniy2022embeddings,shtoff2024functionbasis}. Learnable univariate functions also constitute the edges of Kolmogorov-Arnold networks \citep{liu2025kan}, with specialized implementations such as EfficientKAN \citep{efficientkan2024cao}. These applications illustrate several roles that parametric curves can play within a larger model.

TorchCurves takes the curve itself as the reusable abstraction: a parameterized mapping from a scalar input to a vector output, differentiable with respect to both its input and its coefficients. It provides a common interface for B-spline and Legendre polynomial representations, whose coefficients can be learned directly or supplied by another model. Its contribution is to make these curves convenient to compose with other learning components, without requiring an interpolation problem or prescribing a particular network architecture.

\section{API and applications}\label{sec:applications}
We first present the basic API, and then show use-case examples where it can serve.
\subsection{The API}
The library provides two basic components, \texttt{BSplineCurve} and \texttt{LegendreCurves}, which are one-dimensional curves embedded in a high-dimensional space. Each curve is represented as
\[
  \vc(x) = \sum_{i=1}^n \vv_i B_i(x),
\]
where $B_i$ are the appropriate basis functions, B-Spline or Legendre, and $\vv_1, \ldots, \vv_n$ are its (possibly) learned coefficients. Each component can evaluate a bulk of curves $\vc_1, \ldots, \vc_m$ simultaneously, as shown in Figure \ref{fig:basic_op}. Below is an example:
\begin{pycode}
import torch
import torchcurves as tc

u = torch.rand(4, 3)           # 4 samples with 3 features.
curve = tc.BSplineCurve(
    num_curves=3, dim=2        # 3 curves in 2 dimensions
)
y = curve(u)                   # (4 samples, 3 curves, 2 dimensions)

print(u.shape, "->", y.shape)  # torch.Size([4, 3]) -> torch.Size([4, 5, 2])    
\end{pycode}
Of course, mathemaically inclined readers will say this example is incomplete. And justifiably so - both B-Spline and Legendre polynomial bases require configuration, such as the polynomial degree, the number of knots in case of B-Splines, and others. Moreover, splines and polynomial approximations work in a \emph{compact interval}, which is nowhere to be seen in the above example.

Indeed, we provide the modules with defaults, but they are are fully configurable. Both B-Splines and Legendre polynomials are assumed to approximate inputs on the interval $\mathcal{U} = [-1, 1]$, and can be configured with a mapping for transforming their input to that interval. For example:
\begin{pycode}
    spl_curve = tc.BSplineCurve(
        num_curves=3, dim=2,      # dimensions
        input_map='real.arctan',  # maps `x` to `2 arctan(x) pi`
        num_knots=20,             # 20 uniformly-spaced knots, with not-a-knot endpoints
        degree=5                  # polynomial splines of degree 5
    )

    lgd_curve = tc.LegendreCurve(
        num_curves=3, dim=5 # dimensions
        input_map='arctan', # same as above
        degree=50           # degree-50 Legendre polynomials
    )
\end{pycode}
Mapping functions include the above-mentioned arctangent function, which is also the default, but also $x \to x / \sqrt{1+x^2}$ called \texttt{real.rational} and is inspired by by its use in the differential equations literature on unbounded domains \citep{guo2004legendre,boyd1987spectral,wang2004modified}, and a simple clamping to $\mathcal{U}$ called \texttt{real.clamp}. The prefix \texttt{real} means these mappings are designed for arbitrary real numbers. The library also has \texttt{nonneg.rational} and \texttt{nonneg.arctan} for non-negative numbers. The users can also provide their own, in the form of a Python callable.

Trainable modules are useful, but they do not cover all use-cases the library aims to support. Thus, the library provides three API levels which we shall explain below: (a) trainable modules, as we just seen; (b) basis modules, which cache tensors to avoid their recomputation but have no trainable parameters; (c) a fully functional API, akin to \texttt{torch.nn.functional}.

The fully functional API, as its name suggests, consists of pure functions that have no state. Below is an example:
\begin{pycode}
import torchcurves.functional as tcf

bspl_func = tcf.bspline_curves(
    inputs,   # (B, C) - a batch of input points for each curve
    coefs,    # (C, P, D) coefficients of C B-Spline functions having P coefficients in D dimensions
    knots,    # B-Spline knots. unif. spaced in [-1, 1] with not-a-knot conditions, if unspecified.
    degree=5  # the piecewise-polynomial degree. 3 if unspecified.
)

leg_func = tcf.legendre_curves(
    inputs,   # (B, C) - a batch of input points for each curve
    coefs,    # (C, P, D) coefficients of C legendre polynomials having P coefficinets in D dimensions
    degree=50 # the polynomial degree. We always have P=degree-1.
\end{pycode}
These are useful if the coefficients are not learned directly, but come as an output of another module. Note, that in case of B-Spline functions, the connection between the number of knots and the number of coefficients \texttt{P} must be $\mathtt{P}=|\mathtt{knots}|-\mathtt{degree}+1$. If the knot vector is not specified, the number of uniformly-spaced knots is automatically determined from the degree and the number of coefficients \texttt{P}.

Finally, in between learnable modules and fully functional API we have the basis modules. A basis module is similar to the functional API, but stores immutable buffers, such as the knot vector, for convenience. That way, if a model is moved from CPU to an accelerator, such as a GPU, the buffer automatically moves with it by PyTorch. Below is an example:
\begin{pycode}
import torchcurves as tc

# quadratic splines with 50 B-Spline functions and a rational mapping to [-1, 1]
bspl_basis = tc.BSplineBasis(degree=2, knot_config=50, input_map="real.rational") 

# evaluate a B-Spline function. Note - 50 coefficients for our 50 B-Spline functions.
bspl_func = bspl_basis(
    input,  # (B, C)     - a batch of input points for each curve
    coefs   # (C, 50, D) - the coefficients of C B-Spline functions in D dimensions.
)
\end{pycode}

\subsection{Applications}
One of the first applications that comes to mind is producing embeddings for continuous features, such as height, weight, BMI, the time since the last visit of the user on our website, and others. Imagine we have a model whose inputs are embedding vectors, such as a Factorization Machine, or a Transformer. For discrete, or categorical features - we can just map features to embeddings using a standard embedding table. But for a continuous value, we need a mapping of the value to a vector in the embedding space - and a curve is such a mapping. Here is an example that uses Legendre polynomial basis for such curves:
\begin{pycode}
from torch import nn, cat
import torchcurves as tc

class Net(nn.Module):
    """Composes PyTorch `nn.Embedding` and Legendre curves with another model"""
    def __init__(self, n_discrete, n_continuous, degree, dim):
        super().__init__()
        self.discrete = nn.Embedding(n_discrete, dim)
        self.cont = tc.LegendreCurve(n_continuous, dim, degree)
        self.inner_model = my_super_duper_transformer()

    def forward(self, cat, cont):
        # concatenate all embeddings into one tensor
        embeddings = cat((self.discrete(cat), self.cont(cont)), dim=-2)

        # invoke our desired model on the embeddings
        return self.inner_model(embeddings)
\end{pycode}

Kolmogorov-Arnold networks are also an immediate application - each layer computes many univariate functions of the input, and sums-up their results. Recalling that the output of our curve layers has the shape \texttt{(sample, curve, dimension)}, a KAN layer just applies our curve layer, followed by a sum over the \texttt{curve} dimension. For convenience, we include the \texttt{Sum} module in the library as well. Here is an example:
\begin{pycode}
from torch import nn
import torchcurves as tc

kan = nn.Sequential([
    # layer 1
    tc.BSplineCurve(input_dim, hidden_dim),
    tc.Sum(dim=-2),

    # layer 2
    tc.BSplineCurve(hidden_dim, hidden_dim),
    tc.Sum(dim=-2),

    # layer 3
    tc.BSplineCurve(hidden_dim, output_dim),
    tc.Sum(dim=-2),    
])
\end{pycode}

A less trivial application is that of functional regression - a model whose output is itself a function. In theory, we can think of any model $f(\vx)$ as "functional regression" - just partition $\vx = (\vx_1, \vx_2)$ and define the model as the function $f(\vx_1, \cdot)$. The point is not in the model's architecture, but as the study of the model, including its performance, as such a function. Since the authors come from the online advertising field, one prominent example is fitting the probability of winning an ad auction given auction features and a \emph{non-negative} bid. See \citet{bidshading2020pan,rtb2021zhou}. 

Note that this is a CDF - it must be nondecreasing. A natural choice is using the property of B-Spline functions - if the coefficient vector is non-decreasing, so is the function. So below is such an example - we encode the auction into a non-decreasing vector of coefficients, which in turn serves as the coefficient vector of a B-Spline function of the bid. So below is a model for learning such a probability:
\begin{pycode}
from torch import nn
import torch.nn.functional as F
import torchcurves as tc

class AuctionWinModel(nn.Module):
    def __init__(self):
        super().__init__()
        self.encoder = MySuperBestAuctionEncoder()
        self.spline = tc.BSplineBasis(input_map="nonneg.arctan")

    def forward(self, auction_features, bids):
        # encode auction to a latent vector
        raw = self.encoder(auction_features)

        # convert the latent vector to non-decreasing coefficients:
        #   use first component as is, add cumulative sum of softplus(rest)
        increments = F.softplus(raw[:, 1:])
        coefficients = raw[:, :1] + F.pad(increments.cumsum(-1), (1, 0))

        # non-decreasing spline function of bids to produce P(win) logits.
        return self.spline(bids[None], coefficients[..., None]).flatten()    
\end{pycode}

\section{Algorithms}\label{sec:impl}
Beyond being vectorized, the implementations we use should also be fairly efficient to be useful for machine learning research. Our efficiency goals are modest - we aim for the best we can achieve using pure PyTorch code, without custom kernels. Thus, we implement vectorized versions of well-known algorithms, both for B-Spline and Legendre polynomial evaluation.

The main feature of the B-Spline basis is its sparsity - at any given point, only $\mathtt{degree}+2$ basis functions are non-zero. An efficient algorithm for exploiting the sparsity is the well-known Cox-deBoor method \citep{spline1972cox,spline1972deboor}. The algorithm provides both an efficient sparsity-exploiting way to compute the value \emph{and} the derivative. These, in turn, are used in a custom PyTorch autograd function to facilitate efficient back-propagation through the inputs and the coefficients. The implementation is done in a vectorized manner without iterating on any of the input dimension, but \emph{does} require iterating on the degree. Thus, a spline of degree five does require six Python iterations to compute. 

For Legendre polynomials we have the well-known Clenshaw recursion \citep{clenshaw1955note}. Here, as well, we do not iterate on any input tensor dimensions, but we do iterate on the degree. Since typically good approximation is achieved with high-degree polynomials, higher than B-Splines, this may be less efficient in some situations. For example, a polynomial of degree 50 will require 51 Python for-loop iterations.

In our implementation we do take special care to avoid memory allocation of tensors. We do not claim optimality, but merely a best-effort. Again, our efficiency claim is modest - to be fast enough to be useful.

The library is thoroughly tested, and in particular, the results are compared to SciPy \citep{scipy2020pauli} splines and NumPy \citep{numpy2020harris} Legendre polynomials. Derivatives are validated using internal gradient-checking mechanisms of PyTorch, that compare numerically approximated derivatives to those of back-propagation. Comparisons were done for both 32 and 64 bit floating point numbers.

\section{Experiments}
Our objective here is giving the reader a qualitative understanding of what we mean by "our efficiency claim is modest". To that end, we perform two comparisons: (a) speed of inference-only computation of Legendre polynomials and Splines on CPU, compared to NumPy and SciPy; and (b) training speed of a KAN, compared to EfficientKAN \citep{efficientkan2024cao}, both on CPU and GPU, each implemented in its idiomatic form. Our aim is not to outperform specialized implementations, but to show that the flexibility of treating curves as general-purpose trainable primitives remains computationally practical, even when composing them into higher-level architectures such as KANs.

All measurements were done on a Lenovo P14s Gen-7 laptop, with an Intel Core Ultra 7 356H CPU, and a NVIDIA RTX PRO 500 Blackwell GPU. The experiments were running on Ubuntu 26.04.01, with Python 3.12, NumPy 2.5.4, SciPy 1.18.1, and PyTorch 2.15 (CUDA 13.2). CPU affinity is limited to the first cores 1-4, which are the \emph{performance cores} of that CPU, to both allow PyTorch exploit parallelism and obtain consistent speed measurements.

\begin{figure}[tbhp]
    \centering
    \includegraphics[width=0.95\textwidth]{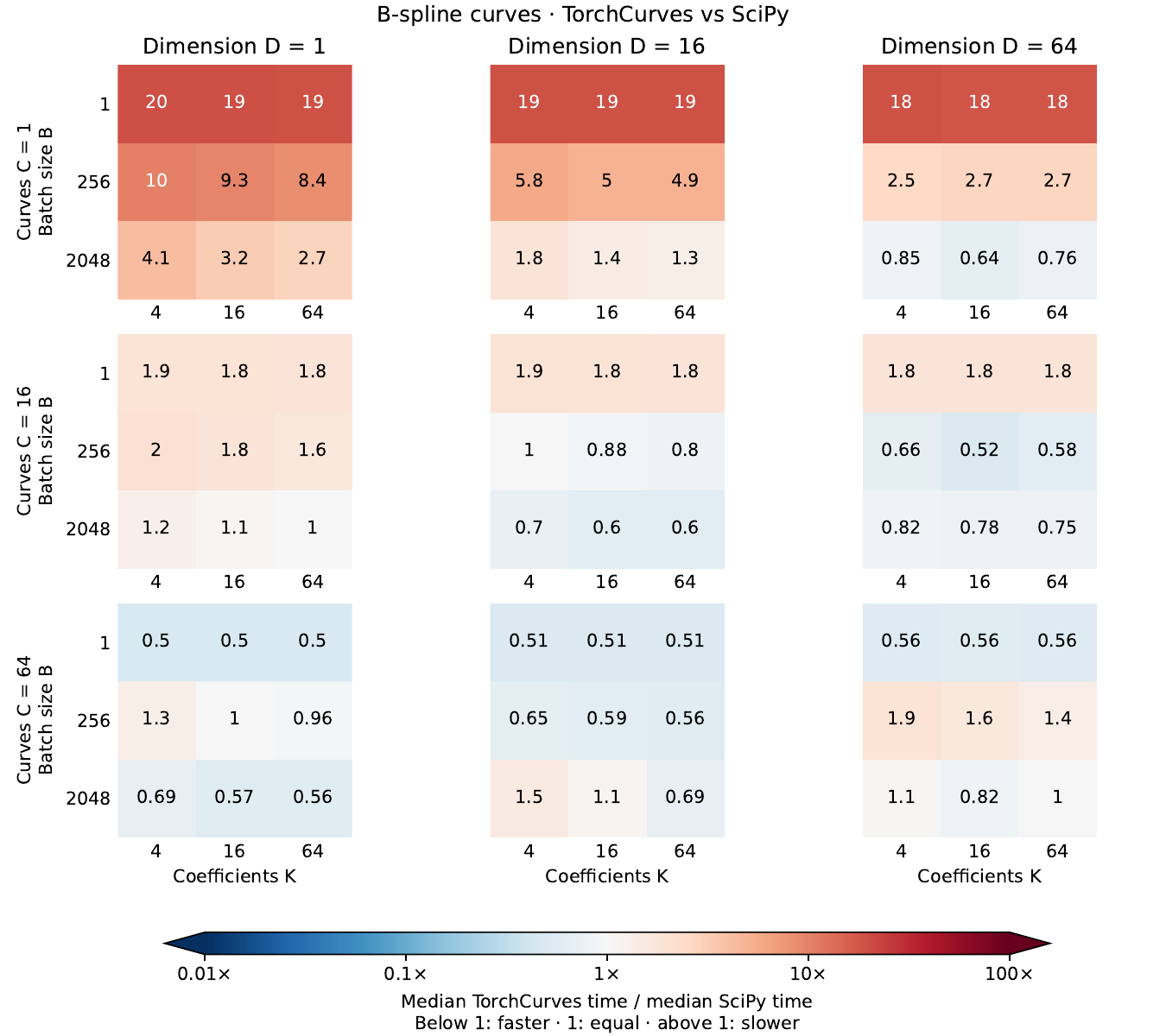}
    \caption{CPU evaluation of independent vector-valued  B-Spline curves. Panel rows show curve count C; panel columns show output dimension D. Within each panel, rows show batch size B and columns show coefficients per scalar output K. Each cell is median TorchCurves time divided by median SciPy time. Values below 1 mean TorchCurves is faster. Medians use five timing measurements. }
    \label{fig:inference_comparison_bspline}
\end{figure}

\begin{figure}[tbhp]
    \centering
    \includegraphics[width=0.95\textwidth]{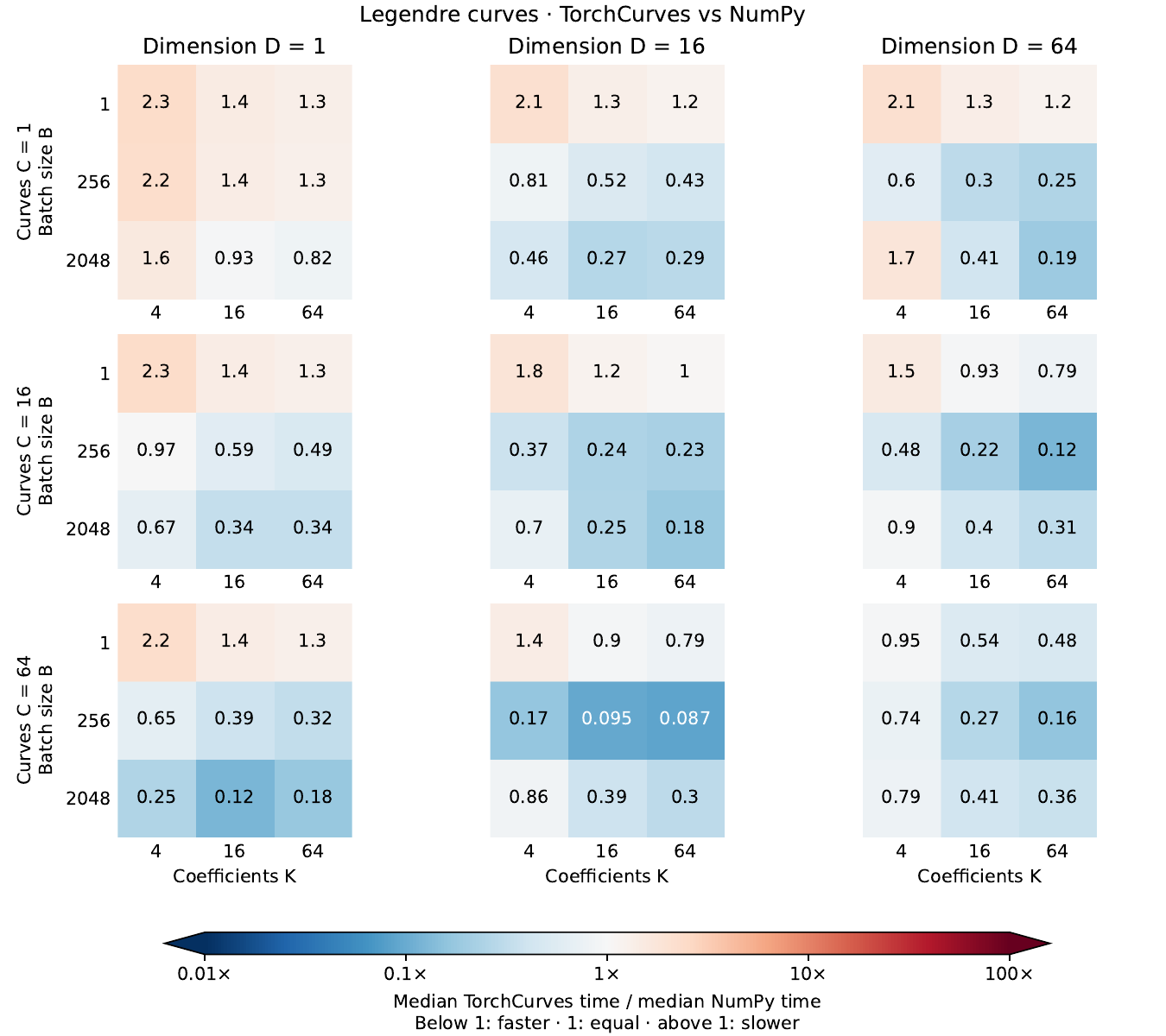}
    \caption{CPU evaluation of independent vector-valued Legendre-polynomial curves. Panel rows show curve count C; panel columns show output dimension D. Within each panel, rows show batch size B and columns show coefficients per scalar output K. Each cell is median TorchCurves time divided by median reference time. Values below 1 mean TorchCurves is faster. Medians use five timing measurements.}
    \label{fig:inference_comparison_legendre}
\end{figure}

We note that both NumPy and SciPy do not provide an autograd mechanism, thus they are by definition ``inference-only'' libraries. Therefore, our comparison in (a) are inference-only in nature - we perform no backpropagation, but only compute the value of the spline / Legendre polynomial functions using each library, and compare the number of milliseconds per call for various coefficient counts, curve dimensions, curve counts, and batch sizes. The results are in Figures \ref{fig:inference_comparison_bspline} and \ref{fig:inference_comparison_legendre}. We can see that \texttt{torchcurves} is slower for small number of curves or small batch sizes, and in the ballpark or faster for larger scales. It is our conjecture that at smaller scales Python overhead wins, whereas for larger scales vectorized PyTorch code lets it exploit multiple cores, while NumPy and SciPy remain serial. In all cases, we perform 

\begin{figure}
    \centering
    \includegraphics[width=0.95\linewidth]{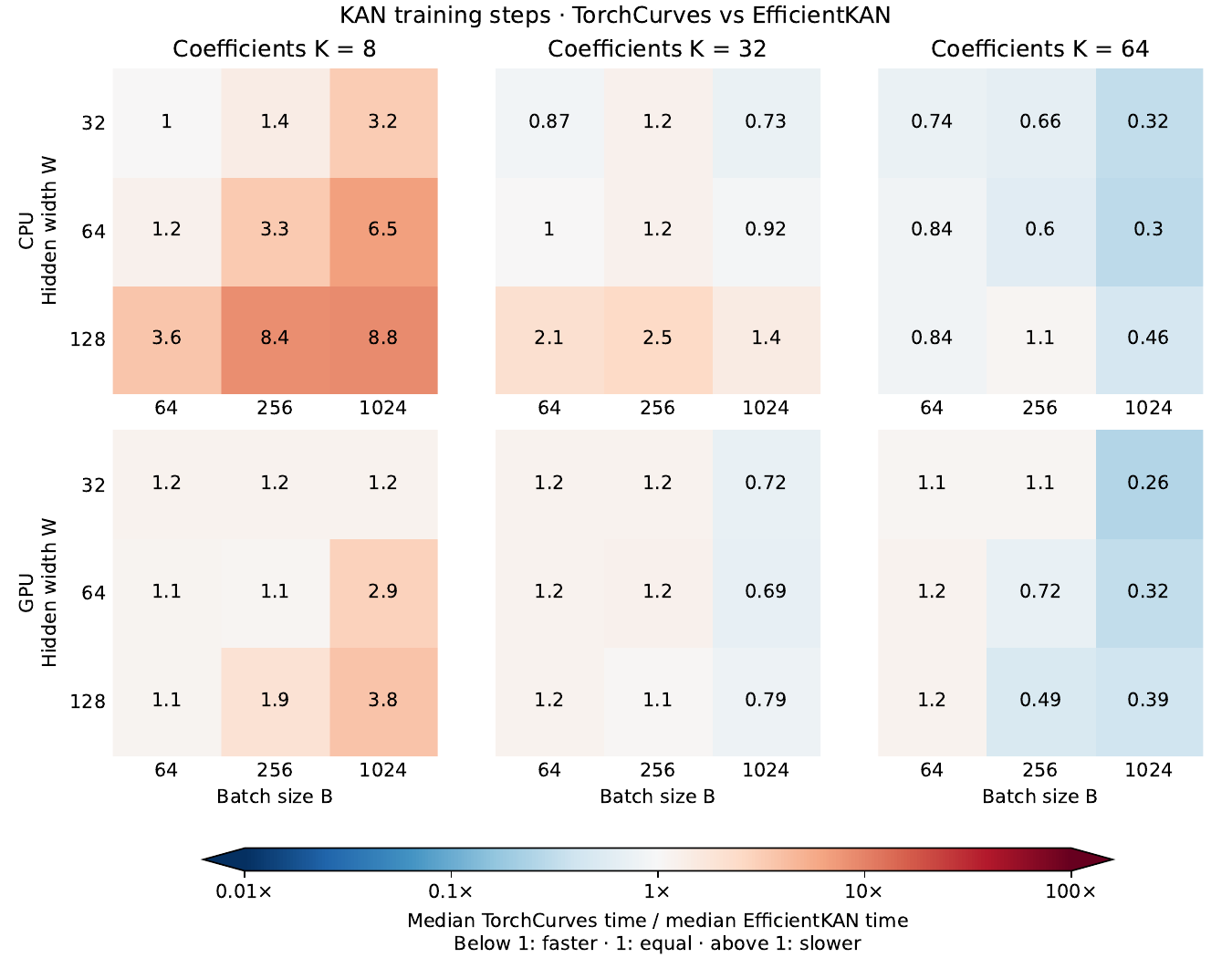}
    \caption{Training-step time ratios for idiomatic [32, W, W, 1] cubic KANs. Panel rows show CPU and GPU; panel columns show coefficient count K. Within each panel, rows show hidden width W and columns show batch size B. Each cell is median TorchCurves time divided by median EfficientKAN time. EfficientKAN uses its KAN model with SiLU branches, trainable spline scaling, extended knots, and no added clamp. Each library initializes its own parameters. Architectures and parameter counts differ. Both use the same generated data, MSE, and Adam with fixed grids. Float32 eager execution.}
    \label{fig:kan_training_time}
\end{figure}

We now turn to KAN training speed comparison. We first note that there could be a different variant - implement a layer that is identical, mathematically, to that of EfficientKAN but implemented as a custom layer using \texttt{torchcurves} functional API. We, on purpose, compare \emph{the} recommended way to build a KAN in both libraries. KANs have two hidden layers, and have the architecture \texttt{[16, W, W, 1]}, meaning, 16 input features, two hidden layers of width \texttt{W}, and a scalar output. We vary the width \texttt{W}, the number of coefficients \texttt{K}, and the training batch size \texttt{B}. Perform 50 warmup steps without measuring time, and then repeat five times training of 100 steps that include clearing gradiens, forward sweep, backward sweep and optimizer step. We divide total train time by 100 to obtain the time per step in each of the five attempts, and then take the median among the attempts to obtain the median time. Results are in Figure \ref{fig:kan_training_time}. We can see that \texttt{torchcurves} is generally in the ballpark of EfficientKAN. Faster for larger batch sizes and higher dimensions, while slower otherwise.

\vskip 0.2in
\bibliography{main.bib}

\end{document}